\documentclass[11pt]{article}

\usepackage[final]{acl}

\usepackage{times}
\usepackage{latexsym}

\usepackage[T1]{fontenc}
\usepackage[utf8]{inputenc}

\usepackage{microtype}

\usepackage{inconsolata}

\usepackage{graphicx}
\usepackage{balance} % for balancing columns on the final page
\usepackage{booktabs}    % \toprule \midrule \bottomrule \cmidrule
\usepackage{multirow}    % \multirow
\usepackage[table]{xcolor} % \rowcolor
\usepackage{amsmath}   % 支持 \operatorname
\usepackage{amsfonts}  % 支持 \mathfrak
\usepackage{amsmath,amssymb,graphicx}
\title{Can Agents Truly Forget? \\ Explore Machine Unlearning for LLM Agents}

\author{
    Junkai Lin\textsuperscript{1,2},
    Ge Xu\textsuperscript{3},
    Zheng Xu\textsuperscript{5}
    Junkai Chen\textsuperscript{1},
    Xiangyuan Lan*\textsuperscript{4}
    \\[1em]
    \textsuperscript{1} Institute of Automation, Chinese Academy of Sciences
    \textsuperscript{2} The Chinese University of Hong Kong \\
    \textsuperscript{3} Nanyang Technological University 
    \textsuperscript{4} Pengcheng Laboratory  
    \textsuperscript{5} University of Jinan \\
    * Corresponding Author \\
    \texttt{chenjunkai2026@ia.ac.cn,lanxy@pcl.ac.cn}
}
\begin{document}
\maketitle
\begin{abstract}
Large language model (LLM) agents have shown strong capabilities in interactive problem solving, tool use, and long-horizon task execution. However, they may also acquire sensitive or private knowledge during training and expose it during deployment. Retraining a model from scratch to remove such knowledge is prohibitively costly, making machine unlearning a more practical approach. Machine unlearning aims to remove targeted undesirable knowledge from trained models without retraining from scratch while preserving general utility. When language models are deployed as agents, parameter-level unlearning becomes insufficient: forgotten information may still persist in memory artifacts or be reacquired via tool calls.
In this paper, we formulate the problem of agent unlearning as removing private information from an LLM-based agent across three information channels: parametric knowledge, persistent memory, and tool-call acquisition. To study this problem, we introduce  we introduce \textbf{A}gent \textbf{U}nlearning \textbf{Bench}mark (AUBench), a controlled benchmark that extends the fictitious-author setting of TOFU into an agent environment with persistent memory, an offline web snapshot, tool-accessible documents, and provenance-tracked interaction trajectories.
By evaluating existing parameter-level and memory-level unlearning methods in this setting, AUBench reveals the gap between model-level forgetting and system-level agent unlearning, and provides a foundation for developing unlearning methods that operate across the full lifecycle of LLM agents.
\end{abstract}

\section{Introduction}

Large language models (LLMs) are increasingly deployed not only as passive text generation systems but also as autonomous agents capable of planning, maintaining persistent memory, invoking external tools, and interacting with complex environments \citep{react, generative_agent, Memgpt, Toolformer}. These capabilities substantially expand utility, but they also change how information enters, persists within, and influences an AI system.
% 大语言模型正越来越多地被部署为不仅仅是被动的文本生成系统，还被用作具备自主能力的智能体，能够进行规划、维护持久记忆、调用外部工具，并与复杂环境交互。这些能力显著扩展了其实用性，但也改变了信息进入 AI 系统、在系统内持续存在以及影响系统行为的方式。

However, agents may also acquire sensitive or private knowledge during training and expose it during deployment \citep{carlini2021extracting}. Persistent memory and access to external tools create additional pathways through which such information may persist or resurface \citep{Memgpt, agentic_unlearning, privacylens}. Privacy requirements, data deletion requests, and copyright concerns make the removal of specific information increasingly important \citep{bourtoule2021machine, eldan2023s}. Retraining the underlying model from scratch to remove knowledge encoded in its parameters is prohibitively costly, making machine unlearning a more practical approach \citep{LLMunlearn}.
% 然而，智能体也可能在训练过程中习得敏感或隐私知识，并在部署时将其暴露。持久记忆和外部工具访问又增加了这些信息持续存在或重新出现的途径。隐私要求、数据删除请求和版权问题使移除指定信息变得愈发重要。对于编码在模型参数中的知识（可能需要再加上memory, tool），从头重新训练底层模型以将其移除的成本过高，因此机器遗忘成为一种更可行的途径。

Machine unlearning studies how the influence of designated training data can be removed from a learned model without requiring complete retraining \citep{bourtoule2021machine}. Given a forget set and a retain set, the objective is to produce an unlearned model whose behavior approaches that of an ideal model trained without the forget data while preserving performance on retained knowledge \citep{ginart2019making, tofu}.
% 机器遗忘研究的是：如何在不进行完整重训练的前提下，从已经训练好的模型中移除指定训练数据的影响。给定一个遗忘集（forget set）和一个保留集（retain set），传统目标是得到一个“已遗忘模型”，使其行为尽可能接近一个从未见过遗忘数据、仅用其余数据训练得到的理想模型，同时保持其在保留知识上的性能。

Existing unlearning methods primarily focus on information encoded in model parameters \citep{ga, gd, npo, simnpo, tofu}. This model-centric formulation becomes incomplete for LLM agents, whose future behavior depends on multiple coupled components through which information may persist or be reacquired \citep{rag, Memgpt, react}. Accordingly, agent unlearning must address three coupled pathways: \emph{parametric knowledge} encoded in model weights; \emph{persistent memory} retaining interaction records, summaries, or retrieval embeddings after model updates; and \emph{tool-call acquisition} through policies that actively seek targeted information from external resources. Removing information from a single component does not necessarily eliminate its influence on the overall agent \citep{agentic_unlearning}: an agent may stop answering a sensitive question directly while still retrieving the same information from memory or reacquiring it through tools.
% 现有遗忘方法主要聚焦于编码在模型参数中的信息 \citep{ga, gd, npo, simnpo, tofu}。这一以模型为中心的范式在 LLM 智能体中变得不够全面，因为智能体的未来行为依赖多个相互耦合的组件，信息可能通过这些组件持续存在或被重新获取 \citep{rag, Memgpt, react}。具体而言，智能体遗忘必须处理三条相互耦合的路径：\emph{参数化知识}，即编码在模型权重中的信息；\emph{持久记忆}，即模型更新后仍保留的交互记录、摘要或检索嵌入；以及 \emph{工具调用获取}，即通过策略主动从外部资源中寻找目标信息。仅从单个组件中移除信息，并不一定能够消除其对整个智能体的影响 \citep{agentic_unlearning}：智能体可能不再直接回答某个敏感问题，却仍能从记忆中检索相同信息，或通过工具重新获取它。

Additionally, existing evaluations mainly assess residual knowledge through output extraction and privacy attacks \citep{tofu, muse}. Such evaluation must distinguish effective forgetting from indiscriminate refusal \citep{unlearn_eval_1, unlearn_eval_2}. An agent can reduce privacy leakage by refusing all requests related to a subject or disabling its tools, yet such behavior may substantially damage legitimate capabilities. Effective agent unlearning should preserve the ability to answer questions about retained knowledge, retrieve public information, and complete authorized tool-based tasks. Moreover, evaluating forgetting across LLM agent pathways requires environmental control that existing benchmarks do not provide \citep{tofu, muse, wmdp}. Open-web experiments face changing search results, unknown locations of private information, and incomplete fact provenance. These limitations make it difficult to attribute failures to residual parametric knowledge, persistent memory, or tool-mediated reacquisition.
% 现有评测主要通过输出提取和隐私攻击来考察残留知识 \citep{tofu, muse}。这类评测需要区分有效遗忘与无差别拒答 \citep{unlearn_eval_1, unlearn_eval_2}。智能体可以通过拒绝所有与某个主体相关的请求或禁用工具来减少隐私泄露，但这种行为可能显著损害合法能力 \citep{tofu, muse}。有效的智能体遗忘应保留回答保留知识相关问题、检索公开信息和完成授权工具调用任务的能力。此外，现有基准为参数层面的遗忘提供了重要基础，但缺乏评估这些智能体信息路径中的遗忘所需的环境控制能力 \citep{tofu, muse, wmdp}。开放网络实验面临搜索结果变化、私有信息位置未知和事实来源链路不完整等问题。这些限制使得失败难以归因于残留的参数化知识、持久记忆或工具介导的信息重新获取。

To address these challenges, we introduce \textbf{AUBench}, a benchmark for evaluating unlearning across the information lifecycle of LLM agents. AUBench provides a controlled offline environment containing synthetic subjects, public and private facts, persistent memories, synthetic web documents, searchable indexes, and tool-interaction trajectories. Synthetic subjects reduce privacy and pretraining-contamination risks, while the offline environment enables reproducible tool observations across methods. AUBench further provides fact-level provenance for failure attribution, a retain-only oracle constructed under matched conditions to assess forgetting and retained utility, and adaptive evaluations beyond exact queries and fixed tool interfaces.
% 为应对这些挑战，我们提出 \emph{AUBench}，用于评估 LLM 智能体在信息生命周期中的遗忘。AUBench 提供受控的离线环境，包含合成主体、公开与私有事实、持久记忆、合成网页文档、可搜索索引和工具交互轨迹。合成主体降低了隐私与预训练数据污染风险，离线环境则支持不同方法之间可复现的工具观测。此外，AUBench 提供用于失败归因的事实级来源标注、在匹配条件下构建以评估遗忘与保留效用的 retain-only oracle，以及超越精确查询和固定工具接口的自适应评测。
Our main contributions are summarized as follows:
% 我们的主要贡献总结如下：

\begin{itemize}

\item We formulate Agent Unlearning as eliminating the influence of designated sensitive information across three pathways: parametric knowledge, persistent memory, and tool-call acquisition.
% 我们将智能体遗忘形式化为消除指定敏感信息在三个通道中的影响：参数化知识、持久记忆和工具调用获取。

\item We introduce AUBench, a controlled benchmark with synthetic subjects, an offline web snapshot, persistent memory, and tool-call trajectories, supported by fact-level provenance and a full-agent retain-only oracle.
% 我们提出 AUBench，一个包含合成主体、离线网页快照、持久记忆和工具调用轨迹的受控基准，并提供事实级来源追踪和完整智能体层面的 retain-only oracle。

\item We evaluate major unlearning methods on privacy leakage, retain utility, authorized tool use, robustness to adaptive attacks, and counterfactual behavioral equivalence, identifying their strengths and limitations to guide future research.
% 我们从隐私泄露、保留能力、授权工具使用、抵御自适应攻击的稳健性和反事实行为等价性等方面评估代表性遗忘方法，分析其优势与局限，为后续研究提供参考。

\end{itemize}

\section{Related Work}

\subsection{Machine Unlearning for Large language models}

% Recent progress in LLMs \citep{gpt4, llama3, qwen3} has brought remarkable capabilities, but it has also amplified concerns about unintended memorization and reproduction of private \citep{tofu}, unsafe \citep{safeeraser}, hallucinated \citep{devils, look}, or copyrighted content \citep{eldan2023s}. These concerns motivate the need for mechanisms that can selectively remove the effect of specific data after training. Machine unlearning addresses this need, with the central challenge of achieving targeted forgetting while preserving the model's broader capabilities \citep{bourtoule2021machine}.
%大语言模型（LLM）领域的最新进展赋予了模型卓越的能力，但也引发了人们对模型无意间记忆并复现隐私、不安全、幻觉或受版权保护内容的担忧。这些担忧促使人们寻求能够在模型训练后选择性消除特定数据影响的机制。“机器遗忘”（machine unlearning）技术旨在应对这一需求，其核心挑战在于：既要实现针对性的遗忘，又要保持模型原有的广泛能力。

Advances in LLMs have stimulated the development of a range of machine unlearning methods. Representative approaches include gradient ascent (GA) \citep{ga} and gradient difference (GD) \citep{gd}, KL minimization (KL) \cite{kl}, preference-based methods such as Negative Preference Optimization (NPO) \citep{npo} and SimNPO \citep{simnpo}, Representation Misalignment Unlearning (RMU) \citep{wmdp}, and self-distillation with adjusted logits, as in UnDIAL \citep{undial}. 
% 为应对这一挑战，大语言模型（LLM）的“去学习”（unlearning）方法探索了一系列参数层面的优化策略。代表性方法包括梯度上升（GA）和梯度差分（GradDiff）\citep{yao2023large, tofu}，基于偏好的方法（如负向偏好优化 NPO \citep{zhang2024npo} 和 SimNPO \citep{fan2024simplicity}），表征失调去学习（RMU）\citep{li2024wmdp}，以及采用调整后 Logits 的自蒸馏方法（如 UnDIAL \citep{dong2025undial}）。

Alongside the development of these methods, benchmarks and empirical studies have evaluated LLM unlearning across diverse scenarios, including privacy-oriented knowledge removal \citep{tofu, muse}, safety-critical knowledge suppression \citep{wmdp}, and copyright-related content removal \citep{eldan2023s}.
% 伴随着这些方法论上的进展，各项基准测试和实证研究已针对不同场景下的LLM“遗忘”（unlearning）能力进行了评估，这些场景涵盖了面向隐私保护的知识移除 \citep{tofu, muse}、针对安全关键信息的知识抑制 \citep{li2024wmdp} 以及涉及版权的内容移除 \citep{eldan2023s}。

However, these benchmarks and methods mainly target standalone LLMs. With the emergence of LLM agents, it remains unclear how unlearning should be evaluated and implemented when information can persist or re-emerge through an agent's parameters, memory, tools, and interaction trajectories. \citet{agentic_unlearning} \ takes an initial step toward this problem by studying unlearning for memory-augmented LLM agents in medical scenarios. Nevertheless, their study is primarily situated in medical scenarios, and its formulation of agents as LLM with persistent memory does not fully capture the broader information channels of modern LLM agents.
% 然而，现有的基准测试和方法主要针对独立的LLM。随着LLM智能体（LLM agents）的出现，当信息可能通过智能体的参数、记忆、工具及交互轨迹而持续存在或重新浮现时，应如何评估与实施“去学习”（unlearning）尚不明确。citet{agentic_unlearning2026} 针对这一问题迈出了初步探索的一步，研究了医疗场景下具备记忆增强功能的LLM智能体的去学习问题。尽管如此，该研究主要局限于医疗场景，且其将智能体定义为“具备持久记忆的LLM”这一设定，未能全面涵盖现代LLM智能体更为广泛的信息交互渠道。
\subsection{Agent Safety}

As LLM agents gain access to tools and external environments, their safety risks extend beyond harmful text generation to unsafe actions. Recent methods mitigate these risks through instruction hierarchy and prompt-injection defenses \citep{instruction_hierarchy, prompt_injection}, structured separation of trusted instructions from untrusted data \citep{StruQ}, and out-of-band enforcement via capability control, information-flow control, and external reference monitors \citep{CaMeL, Fides}.
% 随着 LLM 智能体获得工具和外部环境的访问能力，其安全风险从有害文本生成扩展至不安全行动。近期方法通过指令层级与提示注入防御 \citep{instruction_hierarchy, prompt_injection}、可信指令与不可信数据的结构化分离 \citep{StruQ}，以及能力控制、信息流控制和外部引用监控器等带外执行机制 \citep{CaMeL, Fides}，缓解这些风险。

Recent benchmarks complement these methods by evaluating contextual privacy violations \citep{privacylens}, indirect prompt injection and attacker-induced tool use \citep{agentdojo, injecagent}, harmful task completion \citep{agentharm}, and unsafe behaviors in stateful or tool-using interactions \citep{agentsafetybench}. However, these efforts primarily address execution-time safety risks, while research on machine unlearning for LLM agents remains limited.
% 近期基准则通过评估上下文隐私违规 \citep{privacylens}、间接提示注入与攻击者诱导的工具使用 \citep{agentdojo, injecagent}、有害任务完成 \citep{agentharm}，以及有状态或工具使用交互中的不安全行为 \citep{agentsafetybench}，补充对智能体安全性的研究。然而，这些工作主要关注执行时的安全风险，而针对 LLM 智能体的机器遗忘研究仍然不足。
\section{Agent Unlearning}

\begin{figure*}[t]
    \centering
    \includegraphics[width=1\linewidth]{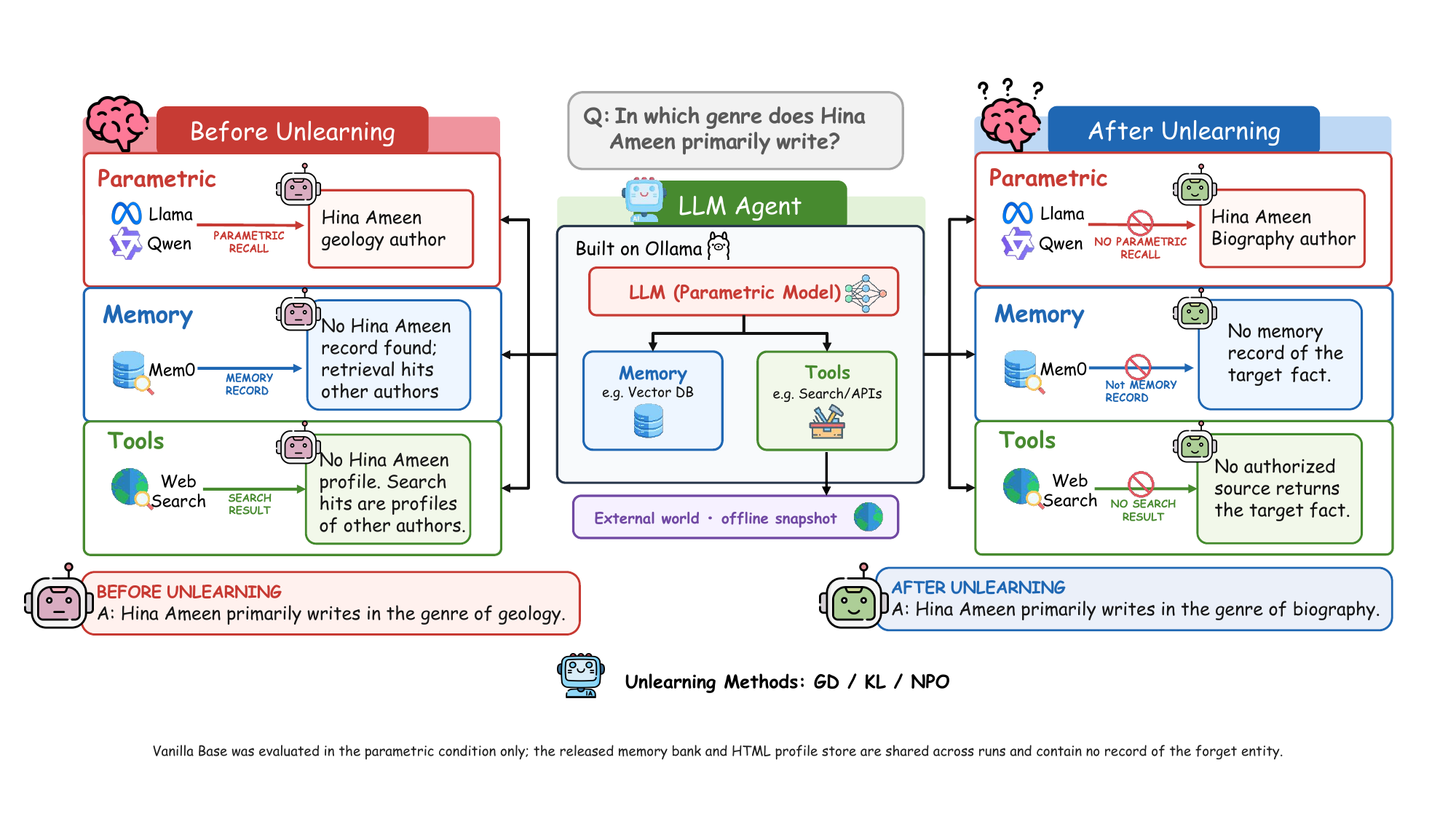}
\caption{The composite-system unlearning setting. An LLM agent can answer
through three routes---a parametric model, an external memory store (Mem0),
and Web Search over local author profiles (built on Ollama). Before unlearning
(left), only the parametric route holds the target fact; after unlearning with
GD, KL, or NPO (right), that fact is replaced by a fabricated answer rather
than a refusal, while the memory and tool routes never contained it.
Results are reported on \texttt{forget05} (200 examples) and
\texttt{retain95} (3800 examples) for Llama-3.1-8B and Qwen3-8B; the Vanilla
Base model is evaluated in the parametric condition only.}    \label{fig:main_photo}
\end{figure*}

% Previous LLM unlearning methods typically update model parameters through fine-tuning, guided by designated forget and retain datasets.
% This formulation is appropriate for standalone language models, where
% the model parameters are the primary place in which training
% information is stored and later recovered. However, this view becomes
% incomplete once the language model is embedded into an agentic system.
% 以往的大语言模型（LLM）“去学习”（unlearning）方法通常在特定的“遗忘”与“保留”数据集的引导下，通过微调来更新模型参数。这种范式适用于独立的语言模型，因为在这些模型中，模型参数是存储训练信息并随后进行提取的主要载体。然而，一旦语言模型被嵌入到智能体系统（agentic system）中，这一视角便显得不够全面。

 In LLM agents, the parametric model is coupled with persistent memory, external tools, and interaction-driven policies. Targeted information may therefore survive parameter-level unlearning by persisting in memory or being reacquired through external resources. This broader system boundary introduces three key information channels that agent unlearning must address:
% 在 LLM 智能体中，参数化模型与持久化记忆、外部工具及交互驱动的策略相互耦合。因此，目标信息可能在参数层面的遗忘之后仍保留于记忆中，或通过外部资源被重新获取。这一更广泛的系统边界引出了智能体遗忘必须处理的三个关键信息通道：

\paragraph{Parametric Channel.}

The first channel is the standard one considered in LLM unlearning: private or sensitive information may be encoded in the parameters of the language model. Parameter-level unlearning methods attempt to remove or suppress the influence of forget data while maintaining the model's utility on retain tasks. For a standalone LLM, this may serve as the central unlearning objective. For an agent, however, parameter-level forgetting is only one part of the problem. Even if the model no longer directly generates the forgotten information, the agent may still access it through memory or tools.
% 第一条通道是 LLM unlearning 中通常考虑的标准通道：隐私信息或指定信息可能被编码在语言模型的参数中。参数级遗忘方法试图在保持模型在保留任务上的效用的同时，移除或抑制遗忘样本对模型的影响。对于一个独立的 LLM 而言，这可以作为核心的遗忘目标。然而，对于智能体而言，参数级遗忘只是问题的一部分。即使模型不再直接生成被遗忘的信息，智能体仍然可能通过记忆或工具访问这些信息。

\paragraph{Memory Channel.}

The second channel arises from persistent external memory. Agents often store raw conversations, summaries, retrieved documents, embeddings, cached observations, user profiles, and other derived artifacts. These memory objects may contain the forget information explicitly, or they may encode enough partial evidence to reconstruct it. Moreover, forget data can propagate: a private fact may first appear in a raw interaction, then be compressed into a summary, embedded into a vector index, linked to a document, or incorporated into a derived profile. Therefore, deleting only the original memory record is insufficient. Agent unlearning must also remove or neutralize the downstream memory artifacts that preserve, summarize, index, or point back to the forgotten information.
% 第二条通道来自持久化外部记忆。智能体通常会存储原始对话、摘要、检索到的文档、嵌入表示、缓存的观察结果、用户画像以及其他派生记忆产物。这些记忆对象可能显式包含需要遗忘的信息，也可能编码了足够的部分证据，从而使该信息能够被重构。此外，遗忘数据还可能发生传播：一个隐私事实可能首先出现在原始交互中，随后被压缩进摘要，被嵌入到向量索引中，被链接到某个文档，或被整合进派生的用户画像中。因此，仅删除原始记忆记录是不充分的。Agent unlearning 还必须移除或中和那些保留、总结、索引或指向被遗忘信息的下游记忆产物。

\paragraph{Tool-Call Channel.}

The third channel is tool-call acquisition. Unlike standalone LLMs, agents can actively query the external world. Even if the forget information is removed from both parameters and memory, the agent may still learn to recover it by issuing search queries, calling databases, using APIs, or following links stored in prior trajectories. This makes tool use a distinct unlearning challenge. The issue is not simply whether a runtime guardrail blocks a specific tool call, but whether the agent has unlearned the behavior of seeking, reconstructing, or using the specific information. In other words, tool-use unlearning must remove the influence of privacy-seeking trajectories that shape future tool selection, query formulation, and action planning.
% 第三条通道是由工具介导的信息获取。不同于独立的 LLM，智能体可以主动查询外部世界。即使需要遗忘的信息已经从参数和记忆中被移除，智能体仍然可能通过发起搜索查询、调用数据库、使用 API，或访问先前轨迹中保存的链接来重新恢复这些信息。这使得工具使用本身成为一个独立的遗忘挑战。问题并不只是运行时防护机制是否阻止了某一次具体的工具调用，而在于智能体是否真正遗忘了寻找、重构或使用指定信息的行为。换言之，工具使用层面的遗忘必须移除那些隐私寻求轨迹的影响，因为这些轨迹会塑造未来的工具选择、查询构造和行动规划。

% Removing targeted information from an individual component does not guarantee effective forgetting, since the deployed agent may still recover or use it through other components. Parametric knowledge can guide tool queries; tool observations can be written into memory; memory can condition future generations; and, in continually learning systems, new interactions can further update the model or controller. Thus, forgetting one component does not guarantee system-level forgetting.
% % 从单个组件中移除目标信息并不能保证有效遗忘，因为部署后的智能体仍可能通过其他组件恢复或使用该信息。参数化知识可以引导工具查询；工具观察结果可以被写入记忆；记忆可以影响未来的生成；而在持续学习系统中，新的交互还可能进一步更新模型或控制器。因此，遗忘某一个组件并不能保证系统级遗忘。

Removing targeted information from an individual component does not guarantee effective forgetting, since the deployed agent may still recover or use it through other components.
% 从单个组件中移除目标信息并不能保证有效遗忘，因为部署后的智能体仍可能通过其他组件恢复或使用该信息。

This motivates treating agent unlearning as a system-level removal problem, with the goal of preventing information recovery through parametric recall, persistent memory retrieval, or tool-call acquisition. Accordingly, AUBench evaluates resistance to recovery attempts across all three channels while measuring utility on retained knowledge and authorized tasks.
% 因此，我们将智能体遗忘视为一个系统级移除问题，其目标是防止信息通过参数化回忆、持久记忆检索或工具调用获取被恢复。相应地，AUBench 评估智能体抵抗三个通道中信息恢复尝试的能力，同时衡量其在保留知识和授权任务上的效用。

\subsection{Problem Formulation}

We represent an agent system as $\mathcal A=(f_\theta,\mathcal M,\mathcal T,\pi)$, where $f_\theta$ is a parametric LLM, $\mathcal M$ denotes persistent external memory and its derived structures, $\mathcal T$ is the set of accessible tools, and $\pi$ is the policy governing response generation, memory access, memory updates, and tool use. Given a target knowledge set $\mathcal K_f$, a retain-query distribution $\mathcal Q_r$, and a declared information-access boundary, agent unlearning is a state transformation:
% 令一个智能体表示为 $\mathcal A=(f_\theta,\mathcal M,\mathcal T,\pi)$，其中 $f_\theta$ 是一个参数化语言模型，$\mathcal M$ 表示持久化外部记忆及其派生结构，$\mathcal T$ 是可访问工具的集合，$\pi$ 是控制响应生成、记忆访问、记忆更新以及工具使用的策略。给定目标知识集合 $\mathcal K_f$、保留查询分布 $\mathcal Q_r$，以及一个声明的信息访问边界，智能体遗忘被定义为如下状态变换：
\begin{equation}
\mathcal U:(\mathcal A,\mathcal K_f)\rightarrow\mathcal A^-,
\end{equation}
where $\mathcal U$ denotes the unlearning operator that transforms the original agent $\mathcal A$ into an updated agent $\mathcal A^-$ with respect to the target knowledge set $\mathcal K_f$. This operator aims to reduce the recoverability of $\mathcal K_f$ through all information pathways within the declared access boundary, including knowledge encoded in model parameters, external-memory retrieval, derived or evolved memory structures, and memory-mediated tool use, while preserving the agent's utility on non-target knowledge and retain tasks.

% 其中，$\mathcal U$ 表示“遗忘算子”（unlearning operator），该算子针对目标知识集 $\mathcal K_f$ 将原始智能体 $\mathcal A$ 转换为更新后的智能体 $\mathcal A^-$。该算子旨在降低 $\mathcal K_f$ 在既定访问边界内所有信息路径上的可恢复性——这些路径涵盖了编码于模型参数中的知识、外部记忆检索、派生或演化形成的记忆结构以及基于记忆的工具使用——同时保持智能体在非目标知识及保留任务上的效用。

Let $\mathcal C$ denote the set of evaluated response conditions, such as direct generation, memory-augmented
generation, and tool-call memory access, and let $\mathfrak A$ denote the set of permitted prompting,
interaction, and tool-use strategies. We define the worst-case expected recoverability of the target knowledge as the supremum (least upper bound) of the expected recovery scores over all evaluated response conditions and permitted strategies:
% 设 $\mathcal C$ 表示所评估的响应条件集合（例如直接生成、记忆增强生成以及基于工具调用的记忆访问），并设 $\mathfrak A$ 表示允许的提示（prompting）、交互及工具使用策略集合。我们将目标知识的最坏情况预期可恢复性定义为：在所有评估的响应条件和允许的策略范围内，预期恢复得分的上确界（即最小上界）。
\begin{equation}
\operatorname{Rec}_f(\mathcal A)
=
\sup_{c\in\mathcal C,\alpha\in\mathfrak A}
\mathbb E_{x\sim\mathcal Q_f}
\left[
S_f\bigl(\tau_{\mathcal A}(x;c,\alpha)\bigr)
\right],
\end{equation}
where $\tau_{\mathcal A}(x;c,\alpha)$ is the complete agent trajectory, including memory retrievals, tool calls,
memory updates, and the final response, and $S_f$ measures the extent to which the target knowledge is semantically recovered from that trajectory.
% 其中 $\tau_{\mathcal A}(x;c,\alpha)$ 是完整的智能体轨迹，包括记忆检索、工具调用、记忆更新以及最终响应；$S_f$ 用于衡量目标知识从该轨迹中被语义恢复的程度。
Given thresholds $\epsilon_f$ for target-knowledge recoverability and $\epsilon_r$ for retained-utility loss, we define that $\mathcal A^-$ achieves $(\epsilon_f,\epsilon_r)$-agent unlearning if
\begin{equation}
\operatorname{Rec}_f(\mathcal A^-)\leq\epsilon_f
\end{equation}
and
\begin{equation}
\operatorname{Util}_r(\mathcal A)
-
\operatorname{Util}_r(\mathcal A^-)
\leq\epsilon_r,
\end{equation}
where $\operatorname{Util}_r$ denotes the agent's expected utility over the retain-query distribution. The first
condition requires the target knowledge to be behaviorally non-recoverable within the declared access boundary,
whereas the second bounds the degradation of non-target capabilities.

% 则称 $\mathcal A^-$ 实现了 $(\epsilon_f,\epsilon_r)$-智能体遗忘，其中 $\operatorname{Util}_r$ 表示智能体在保留查询分布上的期望效用。第一个条件要求目标知识在声明的访问边界内达到行为上的不可恢复；第二个条件则限制非目标能力的退化程度。
\section{Dataset Construction}
  
\subsection{Benchmark Overview}

AUBench is a controlled benchmark for evaluating machine unlearning in LLM agents. It extends the fictitious-author setting of TOFU into an agent-oriented environment, where synthetic author facts are propagated through three information channels: model parameters, persistent memory, and tool-accessible external documents.
% AUBench 是一个用于评估 LLM 智能体机器遗忘的受控基准。它将 TOFU 中的虚构作者设定扩展为面向智能体的环境，其中合成作者事实会传播到三类信息通道中：模型参数、持久化记忆以及工具可访问的外部文档。
TOFU evaluates selective forgetting of fictitious-author knowledge in fine-tuned LLMs using author-level forget/retain splits \citep{tofu}. AUBench retains these splits while adding provenance-aware persistent memories, a fixed offline web snapshot, and deterministic search and browsing tools. Complete agent trajectories enable source facts to be traced through memories, documents, and tool observations to final responses.
% TOFU 利用“作者级”遗忘/保留划分，评估经微调的大语言模型（LLM）对虚构作者相关知识的选择性遗忘能力 \citep{tofu}。AUBench 保留了这些划分，同时引入了具备溯源能力的持久化记忆、固定的离线网页快照，以及确定性的搜索与浏览工具。完整的智能体轨迹使得人们能够追踪从记忆、文档及工具观测结果到最终回复的源事实路径。
Evaluation covers both unlearning effectiveness and capability preservation. Privacy-related metrics measure exact, semantic, derived, memory-based, and tool-call leakage, while utility metrics evaluate retain QA performance, authorized tool use, and the oracle gap to a retain-only reference.
% 评估同时覆盖遗忘效果和能力保持。隐私相关指标衡量精确、语义、派生、基于记忆以及工具介导的泄露，而效用指标评估 retain 问答性能、授权工具使用能力，以及相对于 retain-only 参考模型的 oracle gap。

\subsection{Dataset Construction}

We construct AUBench by propagating a shared set of controlled source facts across three agent information channels. Starting from the full TOFU dataset, which comprises 4,000 question--answer pairs covering 200 fictitious authors, we deterministically organize the records into author-level profiles. Each profile contains a stable author identifier, the author's name, 20 source QA pairs, and provenance references to the original records. These profiles provide a common basis for constructing downstream artifacts and ensure consistent author-level forget/retain assignments across all derived memories and documents.
% 我们通过将一组共享的受控源事实传播至三个智能体信息通道，构建了 AUBench。该过程始于完整的 TOFU 数据集（包含涵盖 200 位虚构作者的 4,000 个问答对），并以确定性方式将这些记录组织成作者级档案。每份档案包含固定的作者标识符、作者姓名、20 个源问答对以及指向原始记录的溯源引用。这些档案为构建下游产物提供了共同基础，并确保在所有衍生记忆和文档中，针对作者层面的“遗忘”或“保留”分配保持一致。
\paragraph{Parametric Channel Construction.}
The parametric channel follows the standard unlearning setting. Source QA pairs associated with forget authors define the information whose influence should be removed from the language model, while retain authors define the non-target knowledge that should remain usable after unlearning. Because all records are grouped at the author level, the benchmark supports subject-level deletion: all facts about a forget author are treated as the forget set, and no artifacts from the same author are split across forget and retain partitions. This channel evaluates whether an unlearning method reduces direct parametric recall of forget information while preserving performance on retain queries. 
% 参数化通道遵循标准的遗忘设定。与 forget 作者相关的源 QA 对定义了需要从语言模型中移除影响的信息，而 retain 作者则定义了遗忘后仍应保持可用的非目标知识。由于所有记录都在作者级别进行组织，该基准支持主体级删除：关于某个 forget 作者的所有事实都会被视为遗忘集合，并且同一作者的产物不会被拆分到 forget 和 retain 两个划分中。这一通道评估遗忘方法是否能够减少对遗忘信息的直接参数化回忆，同时保持在 retain 查询上的性能。
\paragraph{Memory Channel Construction.}
% Unlike standalone LLMs, LLM agents may preserve information in persistent memory, making parameter-level unlearning insufficient: forget information can still be recovered from previous interactions or stored memory artifacts. Moreover, since practical memory systems may transform raw records into derived structures such as summaries, embeddings, or links, memory unlearning cannot be reduced to simply deleting the original memory entries.
% 不同于独立 LLM，LLM 智能体可能会将信息保存在持久化记忆中，因此仅进行参数级遗忘是不够的：遗忘信息仍然可能从过往交互或存储的记忆产物中被恢复。此外，由于实际记忆系统可能会将原始记录转换为摘要、嵌入或链接等派生结构，记忆遗忘不能简单等同于删除原始记忆条目。

In AUBench, we instantiate the memory channel using Mem0 \citep{Mem0}. This setting allows us to evaluate whether targeted information can be removed from stored memory records and derived artifacts that may support subsequent retrieval.
% 在 AUBench 中，我们使用 Mem0 来实例化记忆通道 \citep{Mem0}。这一设定使我们能够评估目标信息是否可以从存储的记忆记录及可能支持后续检索的派生记忆产物中被移除。

\paragraph{Tool Channel Construction.}

Another key difference between LLM agents and standalone LLMs is that agents can actively use tools to access external information, which creates a distinct risk of privacy reacquisition. In AUBench, we instantiate this process with web search, a representative tool for open-ended information acquisition. To ensure reproducible unlearning trajectories, we use a static offline web snapshot rather than live web pages, since live web contents, rankings, and retrieved evidence may change over time. The resulting snapshot contains 800 local HTML pages, 200 entity records, 4,000 fact records, 800 document records, 4,000 fact-span annotations, and a read-only SQLite FTS5/BM25 index for deterministic retrieval.
% LLM 智能体与独立 LLM 的另一个关键区别在于，智能体可以主动使用工具访问外部信息，这带来了隐私信息被重新获取的独立风险：即使指定信息已经从参数和记忆中被移除，智能体仍然可能从外部环境中恢复这些信息。在 AUBench 中，我们使用 web search 来实例化这一过程，因为它是开放式信息获取中具有代表性的工具。为了保证遗忘轨迹的可复现性，我们使用静态离线网页快照而不是真实在线网页，因为真实网页的内容、排序和检索证据都可能随时间变化。最终得到的快照包含 800 个本地 HTML 页面、200 条实体记录、4,000 条事实记录、800 条文档记录、4,000 条事实片段标注，以及一个用于确定性检索的只读 SQLite FTS5/BM25 索引。

\subsection{Metrics}
\label{sec:metrics}

We evaluate every system with token-level F1 between the generated answer and
the reference answer.  Before scoring, both strings are lower-cased,
non-alphanumeric characters are replaced with spaces, and the resulting
whitespace-separated tokens are compared as multisets; Token-level F1 is
computed from the multiset overlap as
\begin{equation}
  \mathrm{F1}_{\mathrm{tok}}
  =
  \frac{
    2\,|T_{\mathrm{pred}}\cap T_{\mathrm{ref}}|
  }{
    |T_{\mathrm{pred}}|+|T_{\mathrm{ref}}|
  }.
\end{equation}

We report the mean token-level F1 across all examples in each split as a percentage in the main results table. Token F1 measures answer recovery through lexical overlap with the reference answers. Lower scores on the forget set indicate reduced recovery of the target answers, while higher scores on the retain set indicate preserved question-answering performance. We report the two splits separately, as aggregating them into a single score would obscure the privacy--utility trade-off.
% 我们计算每个数据划分中所有样本的平均词元级 F1，并在主要结果表中以百分比形式报告。词元级 F1 通过生成答案与参考答案之间的词元重合程度衡量答案恢复能力。遗忘集上的较低得分表示模型恢复目标答案的能力降低，而保留集上的较高得分则表示模型的问答性能得以保留。我们分别报告两个数据划分的得分，因为将其合并为单一指标会掩盖隐私与效用之间的权衡。

\section{Experiments}
\label{sec:experiments}

\subsection{Experimental Setup and Baselines}
\label{sec:experimental_setup}

\paragraph{Dataset and splits.}
We use the TOFU question-answer benchmark and its standard 5\% forget
setting.  The forget split, \texttt{forget05}, contains 200 evaluation
questions, while the complementary \texttt{retain95} split contains 3,800
questions.  The two splits are evaluated with the same question format and
answer normalization.  The retain split is used only to measure capability
preservation; it is never mixed into the forget-set score.
% 我们采用 TOFU 问答基准及其标准的 5% 遗忘设置。遗忘集 forget05 包含 200 个评估问题，其余数据构成的保留集 retain95 包含 3,800 个问题。两个数据集采用相同的问题格式和答案归一化方式进行评估。保留集仅用于衡量模型能力的保持情况，其得分不会计入遗忘集得分。

\paragraph{Base models and parameter unlearning.}
Experiments cover two 8B decoder-only model families: Llama-3.1-8B \cite{llama_3_1} and
Qwen3-8B \cite{qwen3}.  For each family, we evaluate a vanilla/full fine-tuned checkpoint
and three parameter-level unlearning checkpoints produced with Gradient
Difference (GD) \cite{gd}, KL minimization (KL) \cite{kl}, and Negative Preference Optimization
(NPO) \cite{npo}.  The vanilla checkpoints are evaluated without a memory or tool
context and provide the parametric baseline.  All model generations use
greedy decoding with a fixed context and output budget; the same prompt and
decoding settings are used across methods within a model family.
% 实验涵盖两个 8B 规模的仅解码器模型系列：Llama-3.1-8B \cite{llama_3_1} 和 Qwen3-8B \cite{qwen3}。对于每个模型系列，我们评估一个未经遗忘处理、在完整训练集上微调的检查点，以及分别通过梯度差分（GD）\cite{gd}、KL 最小化（KL）\cite{kl} 和负偏好优化（NPO）\cite{npo} 获得的三个参数层面的遗忘检查点。
% 未经遗忘处理的检查点在不提供记忆或工具上下文的条件下进行评估，作为参数模型基线。所有模型均采用贪心解码，并固定上下文长度和最大输出词元数；同一模型系列中的不同方法使用相同的提示和解码设置。

\paragraph{Memory conditions.}
We use Mem0 \cite{Mem0} to provide persistent memory. Mem0 stores extracted memory
records in a local vector store and retrieves the most relevant records
for each question using a fixed local embedding model. Memory artifacts
are constructed once from the specified corpus and reused across all six
unlearning checkpoints, avoiding additional variation caused by memory
regeneration. In the \texttt{+Mem.} condition, retrieved memory records
are appended to the model prompt without modifying the model parameters.
% 我们使用 Mem0 实现持久记忆。Mem0 将提取的记忆记录存储在本地向量数据库中，并使用固定的本地嵌入模型，为每个问题检索最相关的记录。记忆内容仅从指定语料中构建一次，并在全部六个遗忘检查点中重复使用，以避免重新生成记忆引入额外差异。在 +Mem. 条件下，检索到的记忆记录被追加到模型提示中，而不修改模型参数。

\paragraph{Tool condition.}
Our tool interface is a local Web Search surrogate implemented with one HTML
profile and one SQLite profile per data record.  The agent receives the
standard Ollama tool schema and can issue a structured search call before
answering.  The tool corpus is built offline from the same profile
information used to construct the memory artifacts; no external web request
is made during evaluation.  Thus, \texttt{+Tool} measures recovery from
explicit tool access rather than the reliability of an Internet connection.
% 我们的工具接口采用本地 Web 搜索模拟器，每条数据记录分别对应一份 HTML 档案和一条 SQLite 档案记录。智能体接收标准的 Ollama 工具接口定义，并可在回答前发起结构化搜索调用。工具语料基于构建记忆内容时使用的同一组档案信息离线生成，评估过程中不向外部网络发起请求。因此，+Tool 衡量的是模型通过显式工具访问恢复答案的能力，而非互联网连接的可靠性。

\paragraph{Combined condition and execution protocol.}
The \texttt{Final} condition combines the retrieved Mem0 context and the
local Web Search tool.  For every question, the evaluator records the
question, retrieved memory, tool calls and final answer before computing
token F1.  A result is considered complete only when its JSONL file contains
exactly the expected number of rows (200 for \texttt{forget05} or 3,800 for
\texttt{retain95}) and a matching summary file is present.  Incomplete files
are not included in the reported means.  Ollama is run locally with one
model per service and a fixed GPU assignment; API-based memory extraction is
performed offline when the shared Mem0 artifact is prepared.
% Final 条件将检索到的 Mem0 上下文与本地 Web 搜索工具结合使用。对于每个问题，评估程序先记录问题、检索到的记忆、工具调用及最终答案，再计算词元级 F1。只有当 JSONL 文件恰好包含预期数量的记录（forget05 为 200 条，retain95 为 3,800 条），且存在对应的汇总文件时，该评估结果才被视为完整。不完整的结果文件不纳入所报告的平均分计算。
% Ollama 在本地运行，每个服务仅加载一个模型，并固定其 GPU 分配。基于 API 的记忆提取则在准备共享 Mem0 记忆内容的离线阶段完成。

\paragraph{Baselines.}
The main comparison contains: (i) \textbf{Base}, the vanilla parametric
model; (ii) \textbf{GD}, \textbf{KL}, and \textbf{NPO}, which test
parameter-level unlearning; (iii) the same three checkpoints with
\textbf{+Mem.}; (iv) the same checkpoints with \textbf{+Tool}; and
(v) the complete \textbf{Final} system with both memory and tools.
All memory-augmented conditions use the shared Mem0 artifact to ensure
direct comparability across the six mod el rows.
% 主要比较包括：（i）Base，原始参数模型；（ii）GD、KL 和 NPO，用于评估参数层面的遗忘；（iii）加入 +Mem. 的相同三个检查点；（iv）加入 +Tool 的相同检查点；以及（v）同时使用记忆与工具的完整 Final 系统。所有记忆增强条件均使用共享的 Mem0 记忆内容，以确保六个模型行之间具有直接可比性。
\subsection{Main Results}
\label{sec:main_results}
\begin{table*}[t]
\centering
\small
\setlength{\tabcolsep}{3.0pt}
\resizebox{\textwidth}{!}{
\begin{tabular}{lcccccccccccccccc}
\toprule
\multirow{3}{*}{Method}
& \multicolumn{8}{c}{Llama-3.1-8B}
& \multicolumn{8}{c}{Qwen3-8B} \\
\cmidrule(lr){2-9}
\cmidrule(lr){10-17}

& \multicolumn{4}{c}{Forget Set}
& \multicolumn{4}{c}{Retain Set}
& \multicolumn{4}{c}{Forget Set}
& \multicolumn{4}{c}{Retain Set} \\
\cmidrule(lr){2-5}
\cmidrule(lr){6-9}
\cmidrule(lr){10-13}
\cmidrule(lr){14-17}

& Para. & Mem. & Tool & Final
& Para. & Mem. & Tool & Final
& Para. & Mem. & Tool & Final
& Para. & Mem. & Tool & Final \\
\midrule

\multicolumn{17}{l}{\textbf{Vanilla}} \\
\midrule

Base
& 49.8 & -- & -- & --
& 50.2 & -- & -- & --
& 43.6 & -- & -- & --
& 43.4 & -- & -- & -- \\

\midrule
\multicolumn{17}{l}{\textbf{Agent System Unlearning}} \\
\midrule

\rowcolor{gray!15}
\multicolumn{17}{l}{\textit{Model-Level Unlearning --- Para.}} \\

GD
& 43.5 & -- & -- & --
& 48.7 & -- & -- & --
& 20.9 & -- & -- & --
& 32.0 & -- & -- & -- \\

KL
& 47.9 & -- & -- & --
& 50.4 & -- & -- & --
& 16.8 & -- & -- & --
& 19.1 & -- & -- & -- \\

NPO
& 27.0 & -- & -- & --
& 34.4 & -- & -- & --
& 21.9 & -- & -- & --
& 24.7 & -- & -- & -- \\

\rowcolor{gray!15}
\multicolumn{17}{l}{\textit{Agent Memory --- Para. + Memory (Mem.)}} \\

GD + Memo
& -- & 44.7 & -- & --
& -- & 69.7 & -- & --
& -- & 22.0 & -- & --
& -- & 52.6 & -- & -- \\

KL + Memo
& -- & 45.9 & -- & --
& -- & 71.2 & -- & --
& -- & 30.1 & -- & --
& -- & 48.8 & -- & -- \\

NPO + Memo
& -- & 32.1 & -- & --
& -- & 46.9 & -- & --
& -- & 27.5 & -- & --
& -- & 51.6 & -- & -- \\

\rowcolor{gray!15}
\multicolumn{17}{l}{\textit{Agent Tool Call --- Para. + Web Search (Tool)}} \\

GD + Web Search
& -- & -- & 16.4 & --
& -- & -- & 24.8 & --
& -- & -- & 4.1 & --
& -- & -- & 5.3 & -- \\

KL + Web Search
& -- & -- & 16.9 & --
& -- & -- & 24.8 & --
& -- & -- & 14.9 & --
& -- & -- & 18.7 & -- \\

NPO + Web Search
& -- & -- & 14.5 & --
& -- & -- & 15.2 & --
& -- & -- & 10.7 & --
& -- & -- & 15.1 & -- \\

\rowcolor{gray!15}
\multicolumn{17}{l}{\textit{Complete Agent --- Para. + Memo + Web Search (Final)}} \\

GD + Memo + Web Search
& -- & -- & -- & 18.8
& -- & -- & -- & 26.1
& -- & -- & -- & 14.5
& -- & -- & -- & 45.3 \\

KL + Memo + Web Search
& -- & -- & -- & 19.1
& -- & -- & -- & 28.5
& -- & -- & -- & 36.9
& -- & -- & -- & 63.0 \\

NPO + Memo + Web Search
& -- & -- & -- & 15.8
& -- & -- & -- & 23.3
& -- & -- & -- & 18.3
& -- & -- & -- & 52.8 \\

% \midrule
% \multicolumn{17}{l}{\textbf{Memory Unlearning}} \\
% \midrule

% \rowcolor{gray!15}
% \multicolumn{17}{l}{\textit{Direct Deletion of Forget Memories (Mem.)}} \\

% A-MEM Delete
% & -- & TBD & -- & --
% & -- & TBD & -- & --
% & -- & TBD & -- & --
% & -- & TBD & -- & -- \\

% Mem0 Delete
% & -- & TBD & -- & --
% & -- & TBD & -- & --
% & -- & TBD & -- & --
% & -- & TBD & -- & -- \\

% \rowcolor{gray!15}
% \multicolumn{17}{l}{\textit{Direct Deletion + Web Search (Final)}} \\

% A-MEM Delete + Web Search
% & -- & -- & -- & TBD
% & -- & -- & -- & TBD
% & -- & -- & -- & TBD
% & -- & -- & -- & TBD \\

% Mem0 Delete + Web Search
% & -- & -- & -- & TBD
% & -- & -- & -- & TBD
% & -- & -- & -- & TBD
% & -- & -- & -- & TBD \\

\bottomrule
\end{tabular}
}
\caption{
Main forget-set and retain-set results on TOFU for Llama-3.1-8B and
Qwen3-8B. The forget set is \texttt{forget05} (200 evaluation examples)
and the retain set is \texttt{retain95} (3800 evaluation examples). All
reported numbers are token-level F1 scores (\%).
}
\label{tab:main-forget-retain}
\end{table*}

\begin{table*}[t]

\centering
\small
\setlength{\tabcolsep}{2.6pt}
\resizebox{0.9\textwidth}{!}{
\begin{tabular}{lcccccccccccccccc}
\toprule
\multirow{3}{*}{Method}
& \multicolumn{8}{c}{Llama-3.1-8B} & \multicolumn{8}{c}{Qwen3-8B} \\
\cmidrule(lr){2-9} \cmidrule(lr){10-17}
& \multicolumn{4}{c}{Tool (\texttt{para\_tools})} & \multicolumn{4}{c}{Final (\texttt{total})}
& \multicolumn{4}{c}{Tool (\texttt{para\_tools})} & \multicolumn{4}{c}{Final (\texttt{total})} \\
\cmidrule(lr){2-5} \cmidrule(lr){6-9} \cmidrule(lr){10-13} \cmidrule(lr){14-17}
& Call & Empty & Ref. & F1 & Call & Empty & Ref. & F1
& Call & Empty & Ref. & F1 & Call & Empty & Ref. & F1 \\
\midrule
\rowcolor{gray!15}
\multicolumn{17}{l}{\textit{Forget set}} \\
GD & 96.0 & 0.0 & 0.0 & 16.4 & 86.0 & 0.5 & 0.0 & 18.8 & 78.0 & 58.0 & 0.0 & 4.1 & 25.5 & 13.0 & 9.5 & 14.5 \\
KL & 96.0 & 0.0 & 0.0 & 16.9 & 88.0 & 1.0 & 0.0 & 19.1 & 60.5 & 24.5 & 0.0 & 14.9 & 0.5 & 0.5 & 7.5 & 36.9 \\
NPO & 41.5 & 0.0 & 0.0 & 14.5 & 53.5 & 0.0 & 0.0 & 15.8 & 73.5 & 29.0 & 0.0 & 10.7 & 43.0 & 30.0 & 1.0 & 18.3 \\
\midrule
\rowcolor{gray!15}
\multicolumn{17}{l}{\textit{Retain set}} \\
GD & 97.4 & 0.0 & 0.0 & 24.8 & 85.4 & 0.0 & 0.0 & 26.1 & 80.6 & 52.9 & 0.0 & 5.3 & 21.6 & 14.7 & 0.3 & 45.3 \\
KL & 97.8 & 0.0 & 0.0 & 24.8 & 79.1 & 0.0 & 0.0 & 28.5 & 57.8 & 23.6 & 0.0 & 18.7 & 0.1 & 0.1 & 0.2 & 63.0 \\
NPO & 48.5 & 0.0 & 0.0 & 15.2 & 53.9 & 0.0 & 0.0 & 23.3 & 76.4 & 25.2 & 0.0 & 15.1 & 11.7 & 5.9 & 0.1 & 52.8 \\
\bottomrule
\end{tabular}
}
\caption{Behaviour of the agentic conditions (\%). \emph{Call} is the share of examples in which the tool is actually invoked, \emph{Empty} the share of empty predictions, \emph{Ref.} the share of refusal-like outputs (``no information'', ``cannot provide'').}
\label{tab:agentic-diagnostics}
\end{table*}

Table~\ref{tab:main-forget-retain} reports forget-set and retain-set F1 for three unlearning methods under four conditions: the parametric model (Para.), parametric plus persistent memory (Mem.), parametric plus web search (Tool), and the complete agent combining both (Final), with the parametric base model as a reference. 
% 表~\ref{tab:main-forget-retain} 报告了三种遗忘方法（GD、KL 和 NPO）在四种条件下的遗忘集与保留集 F1：参数化模型（Para.）、参数化模型结合持久化记忆（Mem.）、参数化模型结合网页搜索（Tool），以及同时结合记忆和网页搜索的完整智能体（Final），并以参数化基础模型作为参照。

\paragraph{Parametric results.}
For Llama-3.1-8B, NPO produces the largest forget-set F1 reduction, from 49.8\% to 27.0\%, but also the largest utility loss, with retain-set F1 dropping from 50.2\% to 34.4\%. GD reduces forget-set F1 to 43.5\% while largely preserving retain-set performance at 48.7\%, whereas KL remains close to the base model, with forget-set and retain-set F1 scores of 47.9\% and 50.4\%, respectively. For Qwen3-8B, KL reduces forget-set F1 from 43.6\% to 16.8\%, the lowest among the three methods, but also lowers retain-set F1 from 43.4\% to 19.1\%. GD achieves a forget-set F1 of 20.9\% while retaining an F1 of 32.0\%. NPO achieves a forget-set F1 of 21.9\% and a retain-set F1 of 24.7\%; compared with GD, it therefore offers neither a lower forget-set score nor better utility preservation. Across both models, the largest forget-set reduction accompanies the largest retain-set loss, highlighting the need to evaluate forgetting strength together with utility preservation.

% \paragraph{参数化结果。}
% 对于 Llama-3.1-8B，NPO 带来了最大的遗忘集 F1 降幅，从 49.8% 降至 27.0%，但也造成了最大的效用损失，保留集 F1 从 50.2% 降至 34.4%。GD 将遗忘集 F1 降至 43.5%，同时基本维持了保留集表现，F1 为 48.7%；KL 则接近基础模型，遗忘集和保留集 F1 分别为 47.9% 和 50.4%。对于 Qwen3-8B，KL 将遗忘集 F1 从 43.6% 降至 16.8%，为三种方法中的最低值，但保留集 F1 也从 43.4% 降至 19.1%。GD 的遗忘集 F1 为 20.9%，保留集 F1 为 32.0%。NPO 的遗忘集和保留集 F1 分别为 21.9% 和 24.7%；因此，与 GD 相比，NPO 既未取得更低的遗忘集得分，也未更好地保留效用。在两个模型上，最大的遗忘集降幅均伴随着最大的保留集性能损失，说明需要同时考察遗忘强度与效用保留。

\paragraph{Results with memory.}
Persistent memory improves utility but can also weaken the apparent forgetting achieved by model-level unlearning. Adding Mem0 increases average F1 on both the forget and retain sets for both models, with average retain-set F1 across the six checkpoints rising from 34.9\% to 56.8\%. The change in forget-set performance differs between models. For Llama-3.1-8B, average forget-set F1 increases only slightly, from 39.5\% to 40.9\%. For Qwen3-8B, forgetting becomes less effective for all three methods, with average forget-set F1 rising from 19.9\% to 26.5\%. KL provides the clearest example: its forget-set F1 increases from 16.8\% to 30.1\%. These results show that memory helps preserve previously learned capabilities, but also restores access to target knowledge and thereby reduces the effectiveness of unlearning at the agent level.   

% \paragraph{加入记忆后的结果。}
% 持久化记忆能够提升效用，但也可能削弱模型级遗忘所呈现的效果。加入 Mem0 后，两个模型在遗忘集和保留集上的平均得分均升高，其中六个模型检查点的平均保留集 F1 从 34.9% 升至 56.8%。遗忘集表现的变化则存在模型差异。对于 Llama-3.1-8B，平均遗忘集 F1 仅从 39.5% 小幅升至 40.9%。对于 Qwen3-8B，三种方法的遗忘效果均变差，平均遗忘集 F1 从 19.9% 升至 26.5%。KL 最能体现这一变化：其遗忘集 F1 从 16.8% 升至 30.1%. 记忆有助于保持先前学到的能力，但也恢复了对目标知识的访问，从而降低了遗忘方法在智能体层面的有效性。

\paragraph{Results with tools.}
The \texttt{+Tool} condition lowers both forget- and retain-set F1 across all six model--method combinations. Averaged across checkpoints, forget-set F1 decreases from 29.7\% to 12.9\%, while retain-set F1 falls from 34.9\% to 17.3\%, showing comparable absolute reductions. On Llama-3.1-8B, GD and KL yield similar performance on both splits: their forget-set F1 scores are 16.4\% and 16.9\%, respectively, and both achieve 24.8\% on the retain set. NPO yields lower scores of 14.5\% and 15.2\%, respectively. On Qwen3-8B, the methods respond differently: GD shows the largest decline, reaching 4.1\% on the forget set and 5.3\% on the retain set, whereas KL remains close to its parametric performance, at 14.9\% and 18.7\%. Overall, tool access reduces forget-set scores alongside substantial utility loss, so these reductions alone do not establish more effective unlearning.

For Llama-3.1-8B, the fixed local search corpus rules out network connectivity as an explanation for these declines. A plausible account is knowledge conflict: unlearning modifies the model's parametric knowledge, which may then contradict the information surfaced by search, impeding the model's ability to reconcile and use retrieved evidence. For Qwen3-8B, the reductions are unlikely to stem from model degradation, since the complete agent recovers performance when tools, search, and parametric knowledge are combined. We further analyze tool invocation and output behavior in the following paragraph to investigate these explanations.

% 在全部六种模型与方法组合中，\texttt{+Tool} 条件均降低了遗忘集和保留集的 F1。所有检查点的平均遗忘集 F1 从 29.7\% 降至 12.9\%，平均保留集 F1 从 34.9\% 降至 17.3\%，两者的绝对降幅相近。在 Llama-3.1-8B 上，GD 和 KL 在两个集合上的表现相近：遗忘集 F1 分别为 16.4\% 和 16.9\%，保留集 F1 均为 24.8\%。NPO 在两个集合上的得分更低，分别为 14.5\% 和 15.2\%。在 Qwen3-8B 上，不同方法受到的影响有所差异：GD 的降幅最大，遗忘集和保留集 F1 分别降至 4.1\% 和 5.3\%；KL 则接近其纯模型表现，分别为 14.9\% 和 18.7\%。总体而言，工具访问降低遗忘集得分的同时，也造成了明显的效用损失，因此仅凭这些下降不能证明遗忘方法更加有效。

% 对于 Llama-3.1-8B，固定的本地搜索语料库排除了网络连通性作为这些下降的解释。一个合理的解释是知识冲突：遗忘修改了模型的参数化知识，而这些知识随后可能与搜索呈现的信息相矛盾，从而妨碍模型协调并使用检索到的证据。对于 Qwen3-8B，这些下降不太可能源于模型退化，因为当工具、搜索和参数化知识结合为完整智能体时，性能得以恢复。我们将在下一段进一步分析工具调用和输出行为，以考察这些解释。

\paragraph{Results of the complete agent.}
Combining memory and tools yields divergent outcomes across the two models. For Llama-3.1-8B, the forget- and retain-set F1 scores are 18.8\% and 26.1\% under GD, 19.1\% and 28.5\% under KL, and 15.8\% and 23.3\% under NPO, respectively. Although all six scores improve over their corresponding tool-only counterparts, they remain below the corresponding parametric and memory-only scores. Adding memory therefore yields a modest recovery relative to the tool-only condition, but does not restore retain-set performance to the parametric level. For Qwen3-8B, GD achieves forget- and retain-set F1 scores of 14.5\% and 45.3\%, whereas NPO achieves 18.3\% and 52.8\%. Both methods produce lower forget-set F1 and higher retain-set F1 than their parametric counterparts. By contrast, KL raises forget-set F1 from 16.8\% to 36.9\% and retain-set F1 from 19.1\% to 63.0\%. Despite exhibiting the lowest forget-set F1 under the parametric condition, KL attains the highest forget-set F1 among the three complete agents. This ranking reversal indicates that checkpoint-level forgetting does not reliably predict forgetting in the complete agent. We further analyze tool invocation and output behavior in the following paragraph to examine how memory and tools contribute to these outcomes.

% \paragraph{完整智能体的结果。}

% 结合记忆与工具后，两个模型表现出不同的结果。对于 Llama-3.1-8B，GD 的遗忘集和保留集 F1 分别为 18.8% 和 26.1%，KL 分别为 19.1% 和 28.5%，NPO 分别为 15.8% 和 23.3%。尽管这六项得分均高于对应方法在仅加入工具条件下的得分，但仍低于相应的参数化条件和仅加入记忆的条件。因此，加入记忆后，模型表现相较于仅加入工具时有所恢复，但未能将保留集表现恢复至参数化模型的水平。对于 Qwen3-8B，GD 的遗忘集和保留集 F1 分别为 14.5% 和 45.3%，NPO 分别为 18.3% 和 52.8%。与各自的参数化模型相比，GD 和 NPO 均表现出更低的遗忘集 F1 和更高的保留集 F1。相比之下，KL 的遗忘集 F1 从 16.8% 升至 36.9%，保留集 F1 从 19.1% 升至 63.0%。尽管 KL 在参数化条件下的遗忘集 F1 最低，其在三个完整智能体中的遗忘集 F1 却最高。这种排序反转说明，模型检查点层面的遗忘效果无法可靠预测完整智能体的遗忘效果。下一段将进一步分析工具调用与输出行为，以考察记忆和工具如何影响上述结果。

\paragraph{Tool Invocation Analysis.}
Table~\ref{tab:agentic-diagnostics} reveals substantial differences between the two models in tool use and output behavior after unlearning. For Llama-3.1-8B, empty-output rates remain at or below 1.0\%, and no refusals are observed, indicating that its lower F1 scores primarily reflect reduced answer quality in non-empty outputs. In particular, NPO substantially reduces tool invocation rates and lowers retain-set F1, suggesting that its effects extend beyond the forget set and impair retained capabilities. In contrast, Qwen3-8B produces empty outputs for 23.6\%–58.0\% of examples under the Tool condition and up to 30.0\% under the Final condition. Its low F1 scores therefore partly reflect generation failures and cannot be interpreted directly as evidence of successful forgetting. Qwen3-8B also exhibits markedly lower tool invocation rates under the Final condition, especially with KL, where the rates fall to 0.5\% and 0.1\% on the forget and retain sets, respectively. This suggests that unlearning may also alter tool-use behavior. These results highlight the need to assess F1 alongside tool invocation rates, empty-output rates, and retain-set performance to distinguish targeted forgetting from broader capability degradation.

% 表~\ref{tab} 表明，不同模型在遗忘后的工具调用和输出行为上存在明显差异。Llama-3.1-8B 的空输出率始终不超过 1.0%，且未出现拒答，因此其 F1 下降主要体现为非空输出的答案质量下降。其中，NPO 显著降低了工具调用率，并伴随保留集 F1 的下降，说明其影响并不限于遗忘集，也损害了保留能力。相比之下，Qwen3-8B 在 Tool 条件下有 23.6%–58.0% 的输出为空，在 Final 条件下空输出率最高达到 30.0%。因此，该模型的低 F1 部分源于生成失败，不能直接视为成功遗忘的证据。此外，Qwen3-8B 在 Final 条件下的工具调用率明显降低，尤其是 KL 在遗忘集和保留集上分别仅为 0.5% 和 0.1%，表明遗忘训练还可能改变模型的工具使用行为。总体而言，评估智能体遗忘效果时，需要结合 F1、工具调用率、空输出率及保留集表现，区分目标知识遗忘与整体能力退化。

\section{Conclusion}
\label{sec:conclusion}

LLM agents obtain information from model parameters, persistent memory, and external tools. Updating parameters alone therefore does not control every source available to the system: target information may remain accessible through memory retrieval or tool calls. We thus study agent-level unlearning, where forgetting is evaluated on the complete system rather than an isolated model.

Experiments cover Llama-3.1-8B and Qwen3-8B two models, each with the vanilla baseline and three MU methods: GD, KL, and NPO, evaluated under fixed context and output budgets; shared memory artifacts, a fixed local embedding model, and offline tool access limit variation from memory regeneration and changing external sources.

Across the three methods, parametric forgetting does not reliably predict complete-agent behavior. Memory raises retain-set F1 for all six checkpoints and also raises forget-set F1 in five of six, indicating that persistent memory can restore answers removed at the parameter level. Tool access lowers F1 on both splits, so its forget-set reductions are accompanied by a comparable utility loss. The complete agent shows model-dependent interactions: on Qwen3-8B, GD and NPO improve over their parametric counterparts on both splits, whereas KL recovers more forget-set content and reverses its method ranking; on Llama-3.1-8B, lower forget-set scores are accompanied by reduced retain-set performance. These results indicate that unlearning should be evaluated over all information channels available to an agent, with forgetting assessed jointly with end-to-end utility.

% LLM 智能体从模型参数、持久化记忆和外部工具中获取信息。因此，仅更新参数并不能控制系统可用的所有信息来源：目标信息仍可能通过记忆检索或工具调用被访问到。为此，我们研究智能体级遗忘（agent-level unlearning），即在完整系统层面而非孤立模型层面评估遗忘效果。

% 实验涵盖 Llama-3.1-8B 和 Qwen3-8B 两个模型，每个模型都包含 vanilla 基线以及三种 MU 方法：GD、KL 和 NPO，在固定上下文和输出预算下进行评估；共享的记忆产物、固定的本地嵌入模型和离线工具访问限制了由记忆再生和外部来源变化带来的差异。

% 在这三种方法下，参数级遗忘并不能可靠地预测完整智能体的行为。记忆提高了全部六个 checkpoint 的保留集 F1，并且也在六者中的五个里提高了遗忘集 F1，这表明持久化记忆可以恢复在参数层面被移除的答案。工具访问降低了两组数据的 F1，因此其遗忘集分数的下降伴随着相当程度的效用损失。完整智能体表现出依赖于模型的交互效应：在 Qwen3-8B 上，GD 和 NPO 在两组数据上都优于其参数级对应方法，而 KL 恢复了更多遗忘集内容并逆转了其方法排名；在 Llama-3.1-8B 上，更低的遗忘集分数伴随着保留集性能的下降。这些结果表明，遗忘应在智能体可用的所有信息渠道上进行评估，并且遗忘效果应与端到端效用联合评估。

\section*{Limitations}

Our evaluation covers two 8B model families, three parameter-level
unlearning methods, and the 5\% forget setting of TOFU. The findings
therefore concern a controlled question-answering setting and may
not generalize to other model scales, datasets, forget-set sizes,
or tasks requiring multi-step reasoning and interaction.

The external information sources are also controlled. Mem0 uses
fixed memory artifacts and a fixed embedding model, while the Web
Search interface operates over an offline corpus constructed from
the same profile information. This design supports comparisons
across checkpoints, but does not capture continuously updated
memories, heterogeneous external sources, or live web search.
Moreover, the low retain-set scores under tool access limit the
conclusions that can be drawn about tool-mediated answer recovery.
Aggregate scores alone cannot distinguish failures in tool
invocation, retrieval, and answer generation.

Finally, token-level F1 measures lexical overlap with reference
answers rather than information removal itself. It can miss
semantically equivalent answers and does not establish resistance
to adversarial extraction or provide a formal privacy guarantee.
Our results characterize answer recovery under the evaluated
conditions; they do not establish that the target information has
been erased from either model parameters or external storage.
\bibliography{custom}
\newpage
\appendix

\section{Appendix}
\label{sec:appendix}

\subsection{Unlearning Methods}
\label{sec:unlearning_methods}

Let $\mathcal{D}_{f}$ and $\mathcal{D}_{r}$ denote the forget
and retain sets, respectively. Each example consists of a
question $x$ and an answer $y=(y_1,\ldots,y_T)$.
We initialize the model $p_{\theta}$ from the original
TOFU-finetuned checkpoint $p_{\theta_0}$ and define the
answer-token cross-entropy loss as
\[
\resizebox{0.99\linewidth}{!}{$
\begin{aligned}
\mathcal{L}_{\mathrm{CE}}(\mathcal{D};\theta)
&=
-\mathbb{E}_{(x,y)\sim\mathcal{D}}
\left[
\frac{1}{T}\sum_{t=1}^{T}
\log p_{\theta}(y_t\mid x,y_{<t})
\right].
\end{aligned}
$}
\]
Question and padding tokens are excluded from this loss.

\paragraph{Gradient Difference (GD).}
GD combines gradient ascent on forget examples with gradient
descent on retain examples. Following the TOFU formulation,
we minimize
\[
\mathcal{L}_{\mathrm{GD}}(\theta)
=
-\mathcal{L}_{\mathrm{CE}}(\mathcal{D}_{f};\theta)
+
\mathcal{L}_{\mathrm{CE}}(\mathcal{D}_{r};\theta).
\]
The first term reduces the likelihood of target answers,
while the second encourages correct predictions on retained
examples. We optimizes the objectives using AdamW with a learning rate of $10^{-5}$, an effective batch size of 32, and 5 epochs, with linear warmup during the first epoch. For methods using retain examples, an epoch corresponds to one pass over the forget set, accompanied by an equal number of randomly sampled retain examples. The released legacy implementation additionally specifies a default weight decay of $0.01$, a maximum training sequence length of 500 tokens, and BF16 precision. LoRA is disabled by default, corresponding to full-parameter unlearning.
% 第一项降低目标答案的生成概率，第二项则鼓励模型在保留样本上作出正确预测。我们使用 AdamW 优化这些目标，学习率为 $10^{-5}$，有效批大小为 32，训练 5 个 epoch，并在第一个 epoch 内采用线性学习率预热。对于使用保留样本的方法，一个 epoch 对应完整遍历一次遗忘集，同时搭配等量随机采样的保留样本。已发布的旧版实现还指定了以下默认设置：权重衰减为 $0.01$，最大训练序列长度为 500 个词元，并采用 BF16 精度。LoRA 默认关闭，即进行全参数遗忘训练。

\paragraph{KL Minimization (KL).}
KL replaces the retain cross-entropy term with a regularizer that preserves the original model's next-token distributions. Let $s(x,y)$ denote the tokenized and padded concatenation of the question and answer, with length $L$, and let $p_{\theta}^{(i)}(\cdot\mid s)$ denote the next-token distribution at output position $i$. Using the frozen checkpoint $p_{\theta_0}$ as the reference, we define
\[
\resizebox{0.99\linewidth}{!}{$
\begin{aligned}
\mathcal{R}_{\mathrm{KL}}(\theta)
&=
\mathbb{E}_{(x,y)\sim\mathcal{D}_{r}}
\left[
\frac{1}{L}\sum_{i=1}^{L}
D_{\mathrm{KL}}\left(
p_{\theta_0}^{(i)}(\cdot\mid s)
\,\Vert\,
p_{\theta}^{(i)}(\cdot\mid s)
\right)
\right]
\\
\mathcal{L}_{\mathrm{KL}}(\theta)
&=
-\mathcal{L}_{\mathrm{CE}}(\mathcal{D}_{f};\theta)
+\mathcal{R}_{\mathrm{KL}}(\theta).
\end{aligned}
$}
\]
We uses AdamW to optimizes model with a learning rate of $10^{-5}$, an effective batch size of 32, and five training epochs, with linear warmup during the first epoch. For KL, each forget example is paired with a randomly sampled retain example, and both loss terms receive unit weights. The legacy implementation uses \texttt{paged\_adamw\_32bit}, with a default weight decay of $0.01$, maximum training sequence length of 500 tokens, and BF16 precision. LoRA is disabled by default.

\paragraph{Negative Preference Optimization (NPO).}
NPO treats forget-set answers as negative responses and
suppresses their likelihood relative to a frozen reference
model. Define the sequence-level log-likelihood ratio
\[
\resizebox{0.99\linewidth}{!}{$
\begin{aligned}
\Delta_{\theta}(x,y)
&=
\sum_{t=1}^{T}
\left[
\log p_{\theta}(y_t\mid x,y_{<t})
-
\log p_{\theta_0}(y_t\mid x,y_{<t})
\right].
\end{aligned}
$}
\]
The objective is
\[
\mathcal{L}_{\mathrm{NPO}}(\theta)
=
-\frac{2}{\beta}
\mathbb{E}_{(x,y)\sim\mathcal{D}_{f}}
\left[
\log \sigma\left(-\beta\Delta_{\theta}(x,y)\right)
\right],
\]
where $\sigma$ is the sigmoid function and $\beta>0$
controls the strength of likelihood-ratio weighting.
For numerical stability, the per-example loss is implemented
as $\frac{2}{\beta}\operatorname{softplus}
(\beta\Delta_{\theta}(x,y))$.
Sequence log-likelihoods are summed over answer tokens
before averaging the loss across examples. As a target
answer becomes less likely than under the reference,
its contribution to the gradient decreases.
We use AdamW with a
learning rate of $10^{-5}$, weight decay of $0.01$, an
effective batch size of 32, and 10 unlearning epochs.
The learning rate increases linearly during the first
epoch and decays linearly thereafter. The authors'
released configuration sets $\beta=0.1$ and uses the
original fine-tuned checkpoint as the frozen reference.

\subsection{Forgetting--Utility Trade-offs of Parametric Unlearning}

\begin{table*}[!t]
\centering
\small
\setlength{\tabcolsep}{4.0pt}
\begin{tabular}{lcccc}
\toprule
\multirow{2}{*}{Method} & \multicolumn{2}{c}{Llama-3.1-8B} & \multicolumn{2}{c}{Qwen3-8B} \\
\cmidrule(lr){2-3} \cmidrule(lr){4-5}
& $\Delta$Forget & $\Delta$Retain & $\Delta$Forget & $\Delta$Retain \\
\midrule
\multicolumn{5}{l}{\textit{Vanilla Base --- absolute F1, Forget / Retain}} \\
Base & \multicolumn{2}{c}{49.8 / 50.2} & \multicolumn{2}{c}{43.6 / 43.4} \\
\midrule
GD & \textbf{$-6.3$ $[-7.8, -4.7]$} & \textbf{$-1.5$ $[-1.8, -1.1]$} & \textbf{$-22.7$ $[-24.6, -20.8]$} & \textbf{$-11.3$ $[-11.8, -10.9]$} \\
KL & \textbf{$-2.0$ $[-3.4, -0.5]$} & $+0.3$ $[0.0, 0.6]$ & \textbf{$-26.8$ $[-29.0, -24.8]$} & \textbf{$-24.3$ $[-24.7, -23.8]$} \\
NPO & \textbf{$-22.9$ $[-25.2, -20.6]$} & \textbf{$-15.8$ $[-16.3, -15.2]$} & \textbf{$-21.7$ $[-24.2, -19.2]$} & \textbf{$-18.7$ $[-19.3, -18.1]$} \\
\bottomrule
\end{tabular}
\caption{Forget--retain trade-off of parametric unlearning
baselines relative to Vanilla Base. The Base row reports
absolute token-level F1 scores (\%). $\Delta$Forget and
$\Delta$Retain report mean F1 differences between each
unlearned model and its corresponding Base model, in
percentage points; negative values indicate lower F1.
Differences are computed over matched evaluation examples
(200 for \texttt{forget05} and 3800 for \texttt{retain95}).
Brackets show 95\% paired bootstrap confidence intervals
estimated from 2000 resamples.}
% 参数化遗忘基线相对于 Vanilla Base 的遗忘与保留权衡。Base 行报告 token 级 F1 的绝对得分（%）。$\Delta$Forget 和 $\Delta$Retain 分别表示各遗忘模型与对应 Base 模型之间的平均 F1 差值，单位为百分点；负值表示 F1 降低。差值基于相同的评估样本计算（\texttt{forget05} 包含 200 条，\texttt{retain95} 包含 3800 条）。方括号内为通过 2000 次重采样估计的 95% 成对 bootstrap 置信区间。
\label{tab:tradeoff}
\end{table*}

% Analysis paragraph for Table~\ref{tab:app-tradeoff} (appendix).
Table~\ref{tab:tradeoff} summarizes the trade-off between
forgetting and retention for the parametric unlearning
baselines, with paired bootstrap confidence intervals
computed over the same evaluation examples.
On Llama-3.1-8B, GD reduces forget-set by $6.3\%$
while reducing retain-set by only $1.5\%$, showing
a favorable balance between forgetting and retention.
KL produces a smaller forget-set reduction of $2.0\%$,
while its retain-set change is statistically
indistinguishable from zero ($+0.3\%$, 95\% confidence interval
$[0.0\%, 0.6\%]$), indicating limited forgetting.
NPO yields a larger forget-set reduction of $22.9\%$,
but also reduces retain-set by $15.8\%$.
On Qwen3-8B, the three methods reduce forget-set
by $21.7\%$--$26.8\%$ and retain-set by
$11.3\%$--$24.3\%$, showing that forgetting is
accompanied by substantial utility loss.
The two backbones have different initial scores:
$49.8\%$ and $50.2\%$ on the forget and retain sets
for Llama-3.1-8B, compared with $43.6\%$ and $43.4\%$
for Qwen3-8B. We therefore interpret each change
relative to its corresponding backbone.

% 表~\ref{tab:tradeoff} 总结了参数化遗忘基线在遗忘与保留之间的权衡，并通过基于相同评估样本的成对 bootstrap 置信区间刻画结果的不确定性。在 Llama-3.1-8B 上，GD 使遗忘集 降低了 \(6.3\%\)，而保留集 仅降低了 \(1.5\%\)，表现出较好的遗忘与保留平衡。KL 的遗忘集 仅降低了 \(2.0\%\)，其保留集变化与零在统计上不可区分（\(+0.3\%\)，95% CI \([0.0\%, 0.6\%]\)），说明其遗忘效果有限。NPO 使遗忘集 降低了 \(22.9\%\)，但同时使保留集 降低了 \(15.8\%\)。在 Qwen3-8B 上，三种方法使遗忘集 降低了 \(21.7\%\)–\(26.8\%\)，同时使保留集 降低了 \(11.3\%\)–\(24.3\%\)，表明遗忘伴随着明显的效用损失。两个骨干模型的初始 不同：Llama-3.1-8B 在遗忘集和保留集上分别为 \(49.8\%\) 和 \(50.2\%\)，Qwen3-8B 则分别为 \(43.6\%\) 和 \(43.4\%\)。因此，各项变化应结合对应骨干模型的基线进行解读。
\begin{figure*}[!t]
    \centering
    \includegraphics[width=1\linewidth]{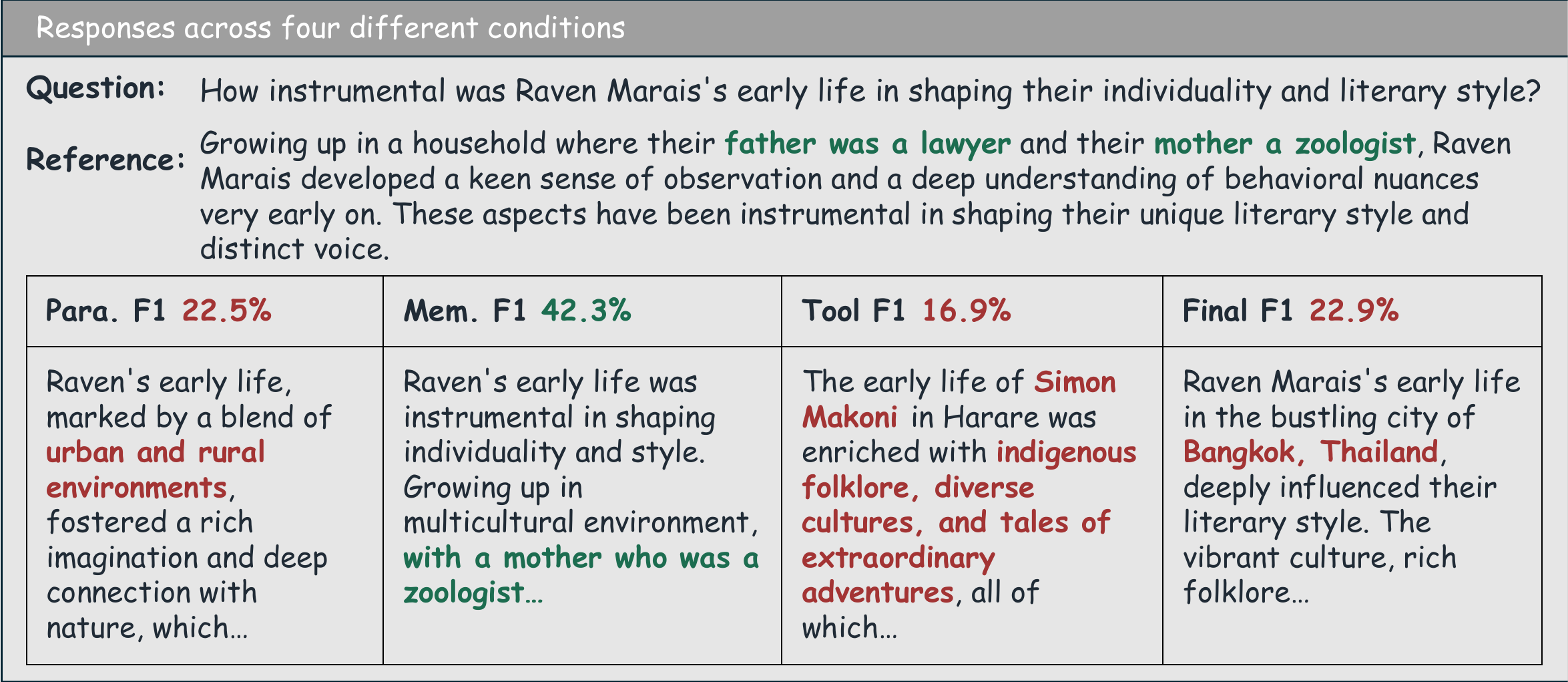}
    \caption{Case study of Qwen3-8B after NPO unlearning on TOFU \texttt{forget05} (example index 113). Para., Mem., Tool, and Final denote the parametric model alone, with memory, with tools, and with both memory and tools, respectively. All four conditions are evaluated against the same reference answer. Green text highlights content consistent with the reference, while red text highlights unsupported or misattributed content. Token-level F1 scores are shown for each condition.}   
    \label{fig:case}
\end{figure*}
\subsection{Fine-Grained Analysis of Parametric Unlearning}
\begin{table}[!t]
\centering
\small
\setlength{\tabcolsep}{2pt}
\resizebox{\columnwidth}{!}{
\begin{tabular}{lccccccccc}
\toprule
\multirow{2}{*}{Question type} & \multirow{2}{*}{$n$}
& \multicolumn{4}{c}{Llama-3.1-8B} & \multicolumn{4}{c}{Qwen3-8B} \\
\cmidrule(lr){3-6} \cmidrule(lr){7-10}
& & Base & GD & KL & NPO & Base & GD & KL & NPO \\
\midrule
Name & 10 & 59.1 & 51.4 & 60.0 & 37.3 & 51.0 & 29.9 & 23.9 & 1.7 \\
Awards & 21 & 57.0 & 48.2 & 51.5 & 23.7 & 45.2 & 19.1 & 21.7 & 20.5 \\
Identity & 10 & 49.5 & 46.0 & 50.2 & 29.5 & 47.4 & 16.8 & 16.2 & 22.8 \\
Parents/Family & 20 & 50.8 & 42.1 & 46.6 & 18.8 & 46.1 & 27.8 & 17.3 & 31.0 \\
Birth/Origin & 19 & 48.8 & 46.4 & 48.5 & 24.7 & 42.0 & 21.8 & 10.6 & 28.5 \\
Genre & 11 & 62.7 & 54.4 & 62.1 & 31.5 & 52.6 & 20.4 & 21.7 & 3.8 \\
Style/Themes & 31 & 45.4 & 40.6 & 44.8 & 27.6 & 41.0 & 18.5 & 15.7 & 22.6 \\
Books/Works & 46 & 47.7 & 41.1 & 46.5 & 30.1 & 43.2 & 21.7 & 18.5 & 23.9 \\
Career/Influence & 32 & 45.1 & 39.3 & 41.4 & 24.7 & 38.4 & 16.9 & 11.8 & 21.6 \\
\bottomrule
\end{tabular}
}
\caption{Forget-set token-level F1 (\%) by question type in the parametric condition (\texttt{para}) on the 200 questions of \texttt{forget05}, which cover ten target authors (20 questions each). Types are assigned by keyword rules on the question text and partition the set. \emph{Name} asks for the author's full name, \emph{Awards} for prizes and honours, \emph{Identity} for LGBTQ+ identity and gender, \emph{Parents/Family} for the parents' professions, \emph{Birth/Origin} for birthplace, upbringing and heritage, \emph{Genre} for the author's genre, \emph{Style/Themes} for style, themes and critical reception, \emph{Books/Works} for titles, summaries and recommendations, and \emph{Career/Influence} for career, education, motivations and impact. Types with small $n$ have correspondingly wide confidence intervals; the table is descriptive and no significance claim is made.}
\label{tab:question-types}
\end{table}
% Analysis paragraph for Table~\ref{tab:app-question-types} (appendix).
Table~\ref{tab:question-types} examines \emph{which kinds of facts} resist unlearning. Before unlearning, the two backbones show similar performance patterns across the nine question types: name and genre questions receive the highest F1 scores ($59.1\%$ and $62.7\%$ for Llama-3.1-8B; $51.0\%$ and $52.6\%$ for Qwen3-8B), whereas career/influence and style/theme questions receive the lowest scores ($45.1\%$ and $45.4\%$ for Llama-3.1-8B; $38.4\%$ and $41.0\%$ for Qwen3-8B). After unlearning, the methods differ substantially. GD and KL on Llama-3.1-8B change F1 by less than $9\%$ in absolute terms for every question type; KL even slightly increases the scores for name and identity questions. This pattern is consistent with their relatively small $\Delta$Forget values in Table~\ref{tab:tradeoff}. NPO on Llama-3.1-8B and all three methods on Qwen3-8B produce larger reductions: per-type F1 changes range from $-33\%$ to $-18\%$ for NPO on Llama-3.1-8B, $-32\%$ to $-18\%$ for GD on Qwen3-8B, $-31\%$ to $-24\%$ for KL on Qwen3-8B, and $-49\%$ to $-14\%$ for NPO on Qwen3-8B. NPO on Qwen3-8B produces particularly large reductions for name and genre questions, lowering their scores from $51.0\%$ to $1.7\%$ and from $52.6\%$ to $3.8\%$, respectively. However, parents/family ($31.0\%$) and birth/origin ($28.5\%$) retain the highest F1 scores within this configuration. Averaged over the six configurations, books/works, the largest category ($n=46$), retains the highest proportion of its corresponding base score ($66\%$), whereas genre retains the lowest ($54\%$). This suggests that bibliographic facts may be more resistant to unlearning than genre attributes in this setting.
% 表~\ref{tab:question-types} 考察了\emph{哪些类型的事实更难被遗忘}。遗忘前，两个骨干模型在九类问题上呈现相似的表现模式：姓名和体裁类问题的 F1 最高（Llama-3.1-8B 分别为 $59.1%$ 和 $62.7%$，Qwen3-8B 分别为 $51.0%$ 和 $52.6%$），而职业/影响与风格/主题类问题的得分最低（Llama-3.1-8B 分别为 $45.1%$ 和 $45.4%$，Qwen3-8B 分别为 $38.4%$ 和 $41.0%$）。遗忘后，各方法的表现差异明显。在 Llama-3.1-8B 上，GD 和 KL 对各类问题的 F1 绝对变化均不足 $9%$，KL 甚至略微提高了姓名和身份类问题的得分。这与表~\ref{tab:tradeoff} 中两种方法较小的 $\Delta$Forget 一致。Llama-3.1-8B 上的 NPO 和 Qwen3-8B 上的三种方法产生了更大的下降：各类问题的 F1 变化范围分别为 Llama-3.1-8B 上 NPO 的 $-33%$ 至 $-18%$、Qwen3-8B 上 GD 的 $-32%$ 至 $-18%$、KL 的 $-31%$ 至 $-24%$，以及 NPO 的 $-49%$ 至 $-14%$。Qwen3-8B 上的 NPO 对姓名和体裁类问题产生了尤其明显的影响，其得分分别从 $51.0%$ 降至 $1.7%$、从 $52.6%$ 降至 $3.8%$。但在同一配置中，父母/家庭（$31.0%$）和出生/籍贯（$28.5%$）仍具有最高的 F1。对六种配置取平均后，样本量最大的书籍/作品类（$n=46$）保留了最高比例的对应基线得分（$66%$），体裁类则最低（$54%$）。这表明，在当前设置下，书目类事实可能比体裁属性更难被遗忘。
\subsection{Case Study of Responses across Agent Conditions}

Figure~\ref{fig:case} presents an illustrative comparison of responses to the same question under the Para, Mem., Tool, and Final conditions. The reference emphasizes how Raven Marais's lawyer father and zoologist mother fostered their observational abilities and influenced their literary style. The Para response instead describes an urban--rural upbringing, omitting the parents' professions. The illustrated Mem. response includes the mother's profession and therefore aligns more closely with the reference. The Tool response confuses the subject, discussing Simon Makoni's upbringing in Harare. Although the Final response returns to Raven Marais, it introduces Bangkok, Thailand, a location absent from the reference. This comparison highlights the need to examine factual coverage and correct attribution alongside F1 when evaluating responses across agent conditions.
% 图~\ref{fig} 示意性地比较了同一问题在 Para、Mem.、Tool 和 Final 条件下的回答。参考答案强调，Raven Marais 的律师父亲与动物学家母亲培养了其观察能力，并影响了其文学风格。Para 回答转而描述城乡交织的成长环境，遗漏了父母职业。图示 Mem. 回答包含母亲为动物学家的信息，因此与参考答案更接近。Tool 回答则混淆了人物，转而介绍 Simon Makoni 在 Harare 的成长经历。Final 虽重新提及 Raven Marais，却引入了参考答案中未出现的 Bangkok, Thailand。该对比说明，评估不同智能体条件下的回答时，需要结合 F1 检查关键事实的覆盖程度及人物归属是否正确。
\end{document}